\pdfoutput=1
\documentclass[10pt, conference]{IEEEtran}

\usepackage{cite}
\ifCLASSINFOpdf
\usepackage{graphicx}
\else
\fi
\usepackage[cmex10]{amsmath}
\usepackage{array}
\usepackage{url}
\usepackage{multirow}
\usepackage{float}

\begin{document}

\title{Information Propagation by Composited Labels in Natural Language Processing}

\author{\IEEEauthorblockN{Takeshi Inagaki}
\IEEEauthorblockA{IBM Japan, Tokyo}
}
\maketitle

\begin{abstract}
In natural language processing (NLP), labeling on regions of text, such as words, sentences and paragraphs, is a basic task. In this paper, label is defined as map between mention of entity in a region on text and context of entity in a broader region on text containing the mention. This definition naturally introduces linkage of entities induced from inclusion relation of regions, and connected entities form a graph representing information flow defined by map. It also enables calculation of information loss through map using entropy, and entropy lost is regarded as distance between two entities over a path on graph.
\end{abstract}


\IEEEpeerreviewmaketitle

\section{Introduction}
NLP is one of active area machine learning is applied and much progress made in recent years. One goal of NLP is extracting knowledge from text documents and represents it in format with explicit structure processable by computers. As a data format of knowledge, RDF and OWL \cite{OWL} on top of it is an example. It is an attempt to represent knowledge in a universal schema, and there are implementations in service such as DBPedia \cite{DBPedia}\cite{DBPedia2}. Knowledge represented in graph structure with RDF schema can be retrieved by a graph query language, SPARQL \cite{SPARQL}. More generally, this kind of approaches are called Knowledge Graph \cite{KG}. Representation of data in minimal pieces (triples) ensures flexibility of reconstructing data in a form user needs at query time. However, its data schema, such as predicates, is to be predefined before processing data, and forming of a graph is relying on matching of subjects and objects as keys to join triples.
Informational source for forming a graph is still an open question, and NLP technologies are expected to extract that information from descriptions in text documents.
In this paper, we focus on geometric information in text document as a source of graph structure.

Output data from NLP is simply labels on text. A label is a name representing category of things described by text. A basic task in NLP is to recognize patterns in sequence of words, and classify them in categories according to meaning they imply. Labeling of small region of text which represents existing object is called entity extraction.  This task determines label of region and boundary of region at same time. A task labeling broader region on text is called as classification. Many studies are on going to improve quality of labeling to replicate human's understanding on text documents by recognizing complex patterns in a sequence of  words.

In this paper, we study information carried by labels delivered by NLP. As mentioned above, labels are bound to regions on text, and structure of a document is designed to represent structure of knowledge. By that, if we have a method to capture document structure information effectively, it will help to reconstruct knowledge from NLP outputs. For this purpose, we employ notion of map, it immediately requires definitions of source set, target set and direction of map. Equivalent class in source set is naturally comes in discussion, and it makes notion of information entropy clear for labeling by NLP models.

\section{Definition of labeling}
In NLP tasks, we consider labeling on various size of regions on text documents. For example, if a text contains a word $Bear$, and if we know he is a dog, we label this region of text with a label $dog$. If $Buddy$ is also a dog, we label this appearance of the word with the label $dog$ too. How can we know they are dogs? The word $Bear$ nor $Buddy$ itself does not provide enough information to determine the label, instead, we need to read text surrounding those words to understand what they mean. When we use machine learning to detect expressions of names of dogs in text documents, it recognizes patterns in surrounding text. One example of machine learning algorithm is BERT \cite{BERT}\cite{BERT2} which recognizes complex patterns within fixed size regions on text. We employ terminology of NLP, we refer regions on text such as $Bear$ and $Buddy$ as mentions, meaning of them implied by their context in text are called entities, $dog$ is a name of entity label in this example. For typical type of entities called named entities, such as person, organization and location, their mentions are small regions on text. However, if we consider entities with more complex context such as regal statements, their mentions will be longer than person names, and to understand their implication, readers should refer broader regions on text including those mentions.

This situation can be summarized as below. On a document text, there is a mention $M$ which is a region $A$ on text, its context is implied by broader region $B$ on text which contains $A$. We can associate this region $B$ to entity $E$ of mention $M$, as
\begin{align}
&A \rightarrow M,  \nonumber \\
&B \rightarrow E,  \nonumber \\
&A \subset B.
\end{align}
To associate mentions and entities each other, we formulate the relation as a map. There are two possibilities, we can consider a map from mention to entity, or a map from entity to mention. To make the map well-defined, for a given element of source set, an element of target set should be determined uniquely. For mentions and entities, there are both of two cases, 1) a single mention is associated to multiple entities, or 2) multiple mentions are associated to a single entity. It means we need to consider both directions of maps.

A map of mention $M$ to entity $E$ happens typically when we consider an entity is a super-class of classes described by mentions. In this case, an entity represents common features of mentions, we define it as
\begin{align}
&f: M \rightarrow E.
\end{align}
For a given mention, a super-class is to be uniquely determined by a map. Note that this does not prevent a class to have has multiple super-classes, but they need to be expressed by multiple maps with different labels. With a single label, an entity is associated to multiple mentions, so it can not be interpreted as mapping from entity to mention, it should be formulated as a map from mention to entity, for example
\begin{align}
dog &\xrightarrow{class} animal, \nonumber \\
cat &\xrightarrow{class} animal,
\end{align}
where $animal$ is the entity for mentions $dog$ and $cat$.

On the other hand, mapping from entity to mention is regarded as description of property of a class, as
\begin{align}
&f': E \rightarrow M.
\end{align}
In this case, a single mention can be referred from multiple entities, for example, $color$ of two $bag$s can be described by a single mention $red$, as
\begin{align}
bag1 &\xrightarrow{color} red, \nonumber \\
bag2 &\xrightarrow{color} red.
\end{align}
An entity $bag1$ can be attributed by multiple maps to mentions, such as
\begin{align}
bag1 &\xrightarrow{color} red, \nonumber \\
bag1 &\xrightarrow{size} small, \nonumber \\
bag1 &\xrightarrow{shape} square.
\end{align}

Note that this direction of map defines direction of information propagation. When $x$ is mapped to $y$ by map $f$, $x$ gains information of $y$. Information flow is inverted direction of map.
This observation is validated in interpretation of mapping as super-class and property. Information of a super-class is applicable to its sub-classes, however, information of sub-classes is not applicable for super-class. Similarly, information of property is applied to its classes. This behavior is consistent with our identification of labeling as map. 

However, distinction between super-class or property is arbitrary. So, we will just refer direction of map only without making distinction between super-class or property. We denote map from mention to entity, in other words, small region to large region containing it, $forward$. Denote map from entity to mention, from large region to small region, $backward$. Discussion in this paper is symmetric on both $forward$ and $backward$ direction map. We can apply observation made on map of a direction to map of other direction.
\begin{figure}[H]
\begin{center}
\includegraphics[width=2.5in]{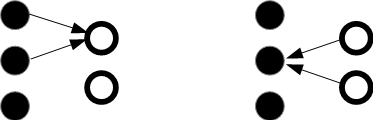}
\end{center}
\caption{\small Black circles represent mentions in text, and white circles are entities represented by mentions. On left hand size, multiple mentions are related to one entity, map can be defined from mentions to entities. We call this mapping $forward$ type. On right hand size, there are multiple entities relating to a mention. The map should be from entities to mentions. We call this mapping $backward$ type.}
\label{Map1}
\end{figure}
There are relations containing both direction of maps. For example, we assume there are two types of bag, $bag1$ and $bag2$. The color of both bags is $black$. It is expressed in text by a mention $black$ and a label $color$ associates it to two entities of $bag1$ and $bag2$ from description in text. In other place on text, it is described as bags of $bag2$ type are owned by a $woman$ and a $man$, a label $owning$ associates them to an entity $bag2$. From this graph, we can know $color$ of a bag a $woman$ $owning$ is $black$.
\begin{figure}[H]
\begin{center}
\includegraphics[width=0.8in]{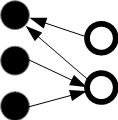}
\end{center}
\caption{\small A black circle at top of left hand side represents mention of color of two types of bags as $black$, two black circles below represent mentions of owners of the same type of bags. A white circle upper side represents type $bag1$ and a circle at lower represents $bag2$. A $bag2$ is either of source and target of map.}
\label{Map2}
\end{figure}
We observed there are two directions of map, and multiple elements of source are mapped in same element of target. This causes loss of information. Actually, NLP is a task of classification. It categorizes elements into smaller number of groups by ignoring details making distinction among them. This information loss is a fundamental nature of NLP. We will address measurement of lost information later.   

\section{Composite map and NLP labels}

NLP task labels various size of regions on text, words, sentences, paragraphs and documents. Small regions are contained in a larger region, and an entity of larger region contains information from small regions in it. So, it is natural to consider a chain of inclusion relations of regions.

By defining label as map, we can convert this problem to a problem of composite map of $f$ and $g$. Consider two maps, $f$ maps mention of region $A$ to an entity $E$ represented by region $B$ which contains region $A$. Region $B$ is regard as a mention $M'$, and it is mapped by $g$ to an entity $E'$ represented by region $C$ which contains $B$, as
\begin{align}
&f: M \rightarrow E, \nonumber \\
&g: M' \rightarrow E',
\end{align}
with inclusion relations
\begin{align}
(A \sim M) \subset (B \sim E \sim M') \subset (C \sim E' \sim M'').
\end{align}
We can naturally consider of composite map as
\begin{align}
g \circ f: M \rightarrow E'.
\end{align}
To simplify notation we identify mention $M$ at step $n$ to entity $E$ at $n-1$
\begin{align}
M_n \simeq E_{n-1}.
\end{align}
We obtain a sequence of mappings in more general form, as
\begin{align}
E_{n-1}  \xrightarrow{f_{n-1}} E_n  \xrightarrow{f_n} E_{n+1}.
\end{align}

An example of a sequence of $forward$ mappings is a categorization problem. A $dog$ and a $cat$ mentioned in text are categorized in $mammal$ in a paragraph, and $mammal$ and $bird$ are categorized in $animal$ in a section. This sequence can be captured by composite maps, as
\begin{align}
&dog \rightarrow mammal \rightarrow animal, \nonumber \\
&cat \rightarrow mammal \rightarrow animal, \nonumber \\
&crow \rightarrow bird \rightarrow animal.
\end{align}

Similarly, text can describe a sequence of $backward$ mappings. Map $color$ and $size$ associate mentions in text $red$ and $small$ to an abstract entity $red.small$. In a broader section, they are aggregated with $shape$ which maps an entity $red.small$ to an entity $red.small.square$, as
\begin{align}
&red.small.square \rightarrow red.small \rightarrow red, \nonumber \\
&red.small.square \rightarrow red.small \rightarrow small, \nonumber \\
&red.small.square \rightarrow square.
\end{align}

We observed NLP label can be regarded as map between mention and entity. By considering regions containing mentions and representing entities, we naturally need to consider sequences of mappings. These sequences form a graph by identification of source and target sets of mappings. This graph is regarded as representing flow of information.
In next section, we will investigate propagation of information by map.

\section{Entropy and Distance}
Our definition of label as map between mention and entity naturally induces information of inclusion relation of regions on text into sequence of map which is regarded as composite map. We can say two entities in different sections in a document are connected, if an entity represented by a mention of large text region containing both sections has maps to two entities in each section. Length of a path connection two entities becomes long if two mentions of entities appear distant locations in a document. It is expected this length indicates distance of two entities in semantic space, it means how distant two entities are corresponding each other.

To make this observation more concrete, we consider information loss by composite map corresponding to a path connecting two entities. Mapping process drops some of information from source set.  We start from a simplest case
\begin{align}
X  \xrightarrow{f} Y.
\end{align}
By map $f$, two or more of elements in source set $X$ is mapped to a same element in target set $Y$. We denote a set of equivalent elements as
\begin{align}
\sim^f = \{a,b \in X| f(a)=f(b)\}.
\end{align}
We can define a quotient set $C$ as below
\begin{align}
C_{f} &= X/\sim^f.
\end{align}
We introduce entropy $S$ as cardinality of set $C$
\begin{align}
S_{f} &= \ln |C_{f}|, \nonumber \\
\Delta S_{f} &= \ln |X| - \ln|C_{f}|.
\end{align}
Entropy is a notion to convert number of states which needs multiplicative operation to calculate number of states of a combined system to additive operation.
$\Delta S_{f}$ is information loss by the map $f$. It can be regarded as distance between $X$ and $Y$. This distance or entropy loss caused by map is accumulated in composite map.
We can restore number of states just by exponentiation of entropy which can be interpreted as probability
\begin{align}
p &= e^{-\Delta S_{f}} = \frac{|C_{f}|}{|X|} \leq 1.
\end{align}
As a next step, we consider composite map
\begin{align}
X  \xrightarrow{f} Y \xrightarrow{g} Z.
\end{align}
A source set of second map $g$ is image of $f$ denoted as $\mathrm{Im}\ f$. It is isomorphic to $C_{f}$ just defined above
\begin{align}
C_{f} &\simeq \mathrm{Im}\ f.
\end{align}
Same as the case of single map, we define a quotient set $C_{g}$ and information loss by composite map $g \circ f$, as
\begin{align}
C_{g} &= \mathrm{Im}\ f/\sim^g, \nonumber \\
\Delta S_{g \circ f} &= \ln |X| - \ln |C_{g}|.
\end{align}

Next, we consider different type of informational connection, where entities are attributed by multiple properties. We suppose we know value of a property, and we would like to know value of other property associated on same entities with knowledge extracted from text by using NLP models. Labels are expected to support this reasoning. How can we capture relations among mentions or entities from analysis of map?
\begin{figure}[H]
\begin{center}
\includegraphics[width=1.2in]{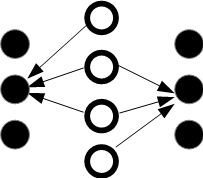}
\end{center}
\caption{\small Black circles represent mentions in text, and white circles are entities represented by mentions. We consider $backward$ map representing properties of entity. Two $backward$ mappings $f$ from center to left, $g$ from center to right are considered. Source sets of entities of map do not need to coincide on both sides.}
\label{Map3}
\end{figure}

In this case, elements of a set of entities $Z$ are mapped to mention $X$ by $f$, and are mapped to $Y$ by $g$, as
\begin{align}
X  \xleftarrow{f} Z \xrightarrow{g} Y.
\end{align}
We would like to know degree of intersection of both map on a set $Z$. We need to work on quotient sets $C_f$ and $C_g$ to count number of colliding elements.
We regard two elements in quotient sets are intersected if elements in two equivalent groups $A \in C_{f}$ and $B \in C_{g}$ are identical or completely included in one of them, $A \subseteq B$.
By denoting number of $A \in C_{f}$ intersecting with $\exists B \in C_{g}$ as $|C_{f} \overrightarrow{\cap} C_{g}|$, we can define entropy from this intersection number, as
\begin{align}
C_{f} &=  Z/\sim^f, \nonumber \\
C_{g} &=  Z/\sim^g, \nonumber \\
S_{f^{-1} \circ g} &= \ln |C_{f} \overrightarrow{\cap} C_{g}|, \nonumber \\
\Delta S_{f^{-1} \circ g} &= \ln |C_{f}| - \ln |C_{f} \overrightarrow{\cap} C_{g}|.
\end{align}
where $e^{-\Delta S_{f^{-1} \circ g}}$ measures ratio of property $f$ propagated to property $g$. If it is $1$, it indicates all values of property $g$ are determined by property $f$.
This provides a systematic way to measure dependency of properties each other.
In an edge case, a set of products is attributed by $size$, $color$ and $price$. In case $price$ is decided just by $size$, does not depend on $color$, we see that intersection number between map $color$ and map $price$ is zero or $S_{f^{-1} \circ g} = -\infty$. It indicates termination of information in this connection.

To validate this observation with a simple example, we consider three sets of target mentions for three types of map, as
\begin{align}
&\mathrm{Im}\ f_{color} = (red, black, blue), \nonumber \\
&\mathrm{Im}\ g_{size} = (small, large), \nonumber \\
&\mathrm{Im}\ h_{price} = (expensive, inexpensive).
\end{align}
As a first example case, we assume existence of rules constraining mappings, as
\begin{align}
g(x)&=small \Rightarrow h(x)=inexpensive, \nonumber \\
g(x)&=large \Rightarrow h(x)=expensive.
\end{align}
Existence of rules restrict possible combination of properties, and we can determine $price$ from other properties $size$ and $color$.
Quotient sets by equivalent classes of three types of map are
\begin{align}
C_f = (&\{ red.large, red.small\},  \nonumber \\
         &\{ back.large, black.small\},  \nonumber \\
         &\{ blue.large, blue.small\}),  \nonumber \\
C_g = (&\{ red.large, black.large, blue.large\},  \nonumber \\
         &\{ red.small, black.small, blue.small\}),  \nonumber \\
C_h = (&\{ red.large, black.large, blue.large\},  \nonumber \\
         &\{ red.small, black.small, blue.small\}).
\end{align}
Rules constrain quotient sets, and intersection numbers among them are,
\begin{align}
&|C_f|=3, |C_{f} \overrightarrow{\cap} C_{h}|=0, \nonumber \\
&|C_g|=2, |C_{g} \overrightarrow{\cap} C_{h}|=2.
\end{align}
This encodes information of rules, in other words, we can restore rules by analyzing intersections on quotient sets with labeled data delivered from text.
In this particular case, it indicates no informational link between $color$ and $price$. On the other hand, entire information is propagated between $size$ and $price$.

We consider another example governed by different rules, where $red$ one is exceptionally $expensive$ regardless of $size$ nor $color$, as
\begin{align}
g(x)&=small \land f(x) \ne red \Rightarrow h(x)=inexpensive, \nonumber \\
g(x)&=large \lor f(x)=red \Rightarrow h(x)=expensive.
\end{align}
This time, quotient set $C_h$ is modified, as
\begin{align}
C_h = (&\{ red.small, red.large, black.large, blue.large\},  \nonumber \\
         &\{ black.small, blue.small\})
\end{align}
We obtain intersection number, as
\begin{align}
&|C_f|=3, |C_{f} \overrightarrow{\cap} C_{h}|=1, \nonumber \\
&|C_g|=2, |C_{g} \overrightarrow{\cap} C_{h}|=1.
\end{align}
In this case, we cannot determine the $price$ from neither of $color$ nor $size$.
To obtain $price$, we need to consider a set $C_f \otimes C_g$ of which each element is a combination of both elements. It consists of every combination of two properties such as $red.large$.
With this, we obtain links from $color \otimes size$ to $price$, as
\begin{align}
&|C_{f} \otimes C_{g}|=6, |(C_{f} \otimes C_{g}) \overrightarrow{\cap} C_{h}|=6.
\end{align}

Finally, we make a remark on entropy of each element in intersecting quotient sets. It has a form of Shannon entropy of information
\begin{align}
S_{entity} &= \frac{1}{|C_{f} \overrightarrow{\cap} C_{g}|} \ln |C_{f} \overrightarrow{\cap} C_{g}|.
\end{align}
This entropy is regarded as measuring amount of information carried by a property value of $f$ against a value of $g$. 
It can be used for evaluation of a relevancy score of property $g$ for $f$. In first example of relation between $size$ and $price$, $size$'s score against $price$ is $\frac{1}{2} \ln 2 = 0.35..$. In second example, relation between $color \otimes size$ and $price$, a combination of $color$ and $size$'s score against $price$ is $\frac{1}{6} \ln 6 = 0.30..$. A property $price$ in first case carries more information of $size$ than of a combination of $color$ and $size$ in second case.

\section{Conclusion}
In this paper, we examined new framework to define NLP labeling in mathematical language. Labels from a graph of which structure is induced from document structure, and it is possible to measure distance between two labeled entities along on a path in the graph. Deductive reasoning by algorithm is one of expected application. 

\bibliographystyle{unsrt}

\begin{thebibliography}{99}

\bibitem{OWL}
Deborah L. McGuinness, Frank van Harmelen
"OWL Web Ontology Language Overview, W3C Recommendation 10 February 2004"
https://www.w3.org/TR/owl-features/

\bibitem{SPARQL}
The W3C SPARQL Working Group
"SPARQL 1.1 Overview, W3C Recommendation 21 March 2013"
https://www.w3.org/TR/2013/REC-sparql11-overview-20130321/

\bibitem{DBPedia}
DBPedia offical site
https://www.dbpedia.org/

\bibitem{DBPedia2}
Christian Bizer, Jens Lehmann, Georgi Kobilarov, Soren Auer, Christian Becker, Richard Cyganiak, Sebastian Hellmann 
"DBpedia - A crystallization point for the Web of Data"
Web Semantics: Science, Services and Agents on the World Wide Web. 7 (3): 154–165. 

\bibitem{KG}
Aidan Hogan, Eva Blomqvist, Michael Cochez, Claudia d'Amato, Gerard de Melo, Claudio Gutierrez, José Emilio Labra Gayo, Sabrina Kirrane, Sebastian Neumaier, Axel Polleres, Roberto Navigli, Axel-Cyrille Ngonga Ngomo, Sabbir M. Rashid, Anisa Rula, Lukas Schmelzeisen, Juan Sequeda, Steffen Staab, Antoine Zimmermann
"Knowledge Graphs"
ACM Computing Surveys. 54 (4): 1–37. arXiv:2003.02320

\bibitem{BERT}
Jacob Devlin, Ming-Wei Chang, Kenton Lee, Kristina Toutanova
"BERT: Pre-training of Deep Bidirectional Transformers for Language Understanding"
arXiv:1810.04805 

\bibitem{BERT2}
Ashish Vaswani, Noam Shazeer, Niki Parmar, Jakob Uszkoreit, Llion Jones, Aidan N. Gomez, Lukasz Kaiser, Illia Polosukhin
"Attention Is All You Need"
arXiv:1706.03762

\end{thebibliography}

\end{document}